\documentclass[runningheads]{llncs}
\usepackage{graphicx}
\usepackage{multirow}

\usepackage{url}

\begin{document}
\title{Refining Diagnosis Paths for Medical Diagnosis based on an Augmented Knowledge Graph}
\titlerunning{Refining Diagnosis Paths 
for Medical Diagnosis}

\author{Niclas Heilig\inst{1,2} \and Jan Kirchhoff\inst{1} \and Florian Stumpe\inst{1} \and Joan Plepi\inst{3} \and Lucie Flek\inst{3} \and Heiko Paulheim\inst{2}\orcidID{0000-0003-4386-8195}}

\authorrunning{N. Heilig et al.}

\institute{medicalvalues GmbH, Karlsruhe, Germany \\\email{\{niclas.heilig,jan.kirchhoff,florian.stumpe\}@medicalvalues.de}\and University of Mannheim, Mannheim, Germany\\\email{heiko@informatik.uni-mannheim.de} \and Philipps-Universität Marburg, Marburg, Germany\\\email{\{joan.plepi,lucie.flek\}@uni-marburg.de}
}

\maketitle              

\setcounter{footnote}{0}
\begin{abstract}
Medical diagnosis is the process of making a prediction of the disease a patient is likely to have, given a set of symptoms and observations. This requires extensive expert knowledge, in particular when covering a large variety of diseases. 
Such knowledge can be coded in a knowledge graph -- encompassing diseases, symptoms, and diagnosis paths.
Since both the knowledge itself and its encoding can be incomplete, refining the knowledge graph with additional information helps physicians making better predictions.
At the same time, for deployment in a hospital, the diagnosis must be explainable and transparent.
In this paper, we present an approach using \emph{diagnosis paths} in a medical knowledge graph.
We show that those graphs can be refined using latent representations with RDF2vec, while the final diagnosis is still made in an explainable way.
Using both an intrinsic as well as an expert-based evaluation, we show that the embedding-based prediction approach is beneficial for refining the graph with additional valid conditions.

\end{abstract}

\section{Introduction}
Medical diagnosis is defined as the process of predicting the disease a patient is likely to have, given a set of symptoms and observations. This process is often not a one shot endeavor, but can involve different steps, such as examinations and the collection of additional evidence (e.g., through blood pictures, X-ray imaging, etc.). We call these processes \emph{diagnosis paths}. In order to arrive at the correct diagnosis, a physician needs to know those paths, collect the required evidence and combine the information about symptoms and observations.

The information on the different symptoms, observations and diseases can be stored in a \emph{knowledge graph} \cite{hogan2021knowledge}. Diagnosis paths can also be formalized in such a knowledge graph, and be used to reason about a patient's diagnosis, just like a medical expert following that path. The \texttt{medicalvalues} knowledge graph encodes such knowledge on more than 380 diseases, together with the corresponding diagnosis paths. Currently, the expert system based on this knowledge graph, offered by the company \texttt{medicalvalues}, is in use in one large laboratory chain (4,000 employees) and also in different pilot stages in leading German university hospitals, including the university hospital in Mannheim (5,000 employees).

In the medical field, transparency, accountability and determinism of medical diagnosis systems are a key requirement for building systems that are accepted by medical experts. The \texttt{medicalvalues} knowledge graph comes with a reasoning system which uses the diagnosis paths to not only provide an expert with suggested diagnoses, but also with an explanation why it came to the conclusion.


The diagnosis paths in the knowledge graph are valid, but usually not complete, and may miss some alternatives and rare side conditions. 
In this paper, we show how patient data can help training a machine learning model to complete the diagnosis paths in the graph. To that end, we explore the use of knowledge graph embeddings \cite{portisch2021embedding,wang2017knowledge} to predict additional edges in the graph.

While it is also possible to utilize knowledge graph embeddings to directly infer a diagnosis, this would be at odds with the aim of providing explainable diagnoses. In contrast, we follow a \emph{knowledge graph refinement} approach \cite{paulheim2017knowledge}, focusing on augmenting the existing diagnosis paths (which then, in turn, provide explainable diagnoses) instead of predicting diseases directly. 
The refined diagnosis paths are evaluated for the validity by medical experts.

The rest of this paper is structured as follows. We review related work in section~\ref{sec:related}. We introduce the \texttt{medicalvalues} knowledge graph in section~\ref{sec:knowledgegraph}, and outline our refinement approach in section~\ref{sec:refinement}, followed by an evaluation in section~\ref{sec:evaluation}. We conclude with a summary and an outlook on future work.

\section{Related Work}
\label{sec:related}
The biomedical domain has been one of the earliest and most vivid adopters of semantic web technologies, ontologies, and knowledge graphs. The Linked Open Data cloud\footnote{\url{https://lod-cloud.net/}} depicts a large subset of linked open datasets coming from the life sciences domain \cite{schmachtenberg2014adoption}, most notably BioPortal~\cite{noy2009bioportal} and Bio2RDF~\cite{belleau2008bio2rdf}. Moreover, a recent survey on domain-specific knowledge graphs showed a wide adoption of knowledge graphs in the healthcare domain.~\cite{abu2021domain}

While knowledge graphs can be used for various purposes in the medical domain, such as the exploration of scientific literature, the focus of our work is on medical diagnosis. Similar approaches are discussed in \cite{chen2019robustly,ernst2014knowlife,rotmensch2017learning,wang2017pdd}, where knowledge graphs of symptoms and diseases are constructed from electronic medical records (EMR), medical literature, and/or other sources by means of relation extraction. However, while those approaches build a bipartite graph of symptoms and diseases, the \texttt{medicalvalues} knowledge graph uses more complex diagnosis pathways and rules (see below).

Some approaches have been proposed in the recent years which cover particular subfields of medicine \cite{liu2016hkdp} or diseases \cite{chai2020diagnosis}. In contrast to those works, the \texttt{medicalvalues} knowledge graph aims at covering a larger variety of diseases.

When it comes to refining existing medical knowledge graphs, the combination of knowledge graph embeddings and machine learning models, similarly to this paper, has been utilized in the past, e.g., for predicting drug-drug interaction \cite{celebi2018evaluation,karim2019drug}, gene-disease interaction \cite{nunes2021predicting}, or other tasks, such as protein-protein interaction, protein function similarity, protein sequence similarity, and phenotype-based gene similarity \cite{sousa2021supervised}. Unlike the work presented in this paper, those approaches mostly use a single in-domain knowledge graph for their predictions, while we present a method using an integrated augmented knowledge graph incorporating numerous types of information.

\section{The \texttt{medicalvalues} Knowledge Graph}
\label{sec:knowledgegraph}
The \texttt{medicalvalues} GmbH\footnote{\url{https://medicalvalues.de/}} is a startup company developing software systems to provide decision support in hospitals. This software helps physicians to assess the risk of suffering from a disease and to decide about further examination steps. All decisions proposed by the software are substantiated with medical guidelines curated by medical experts. The disease information is modeled in a knowledge graph. Medical experts working at \texttt{medicalvalues} use the internal graph editor to encode expert knowledge, extracted from widely-accepted medical sources, in the form of rules for medical diagnosis. 

The customers of \texttt{medicalvalues} are hospitals and laboratory providers. Therefore, integrations into clinic information systems (CIS) and laboratory information systems (LIS) are built. As the laboratory parameters are comparatively easy to analyze by machines, the medical focus lies on the evaluation of laboratory properties. Also, the current medical procedures show that it is possible to improve the laboratory parameters analysis.

In preliminary studies with medical experts, different representations of knowledge graphs (e.g., RDF graphs, labeled property graphs) were explored. Ultimately, labeled property graphs were found to be the most intuitive and usable and were therefore chosen as a representation mechanism for the \texttt{medicalvalues} knowledge graph. The graph is stored in a relational Postgres database, and the user interface for the medical experts is provided as a web application. 

To identify the medical risk factors which can increase the risk to suffer from a disease, the \texttt{medicalvalues} knowledge graph reuses widely accepted coding systems. One of them is the International Classification of Diseases and Related Health Problems (ICD). The ICD system is maintained by the World Health Organization (WHO) and contains codes to exactly identify what a patient suffers from \cite{world1992icd}. The diseases in the \texttt{medicalvalues} knowledge graph are associated with an ICD code, e.g., for epidemiological research or the billing process in hospitals. Furthermore, \texttt{medicalvalues} uses also the SNOMED coding system \cite{snomedManual} to identify diseases, findings, and imaging results. In contrast to ICD, SNOMED is commonly used to describe the medical inputs to draw conclusions on possible diseases. For laboratory parameters, the Logical Observation Identifiers and Codes (LOINC) specialized on the identification of laboratory measurements, provides the identifiers in the graph \cite{loinc-304}. Using common coding systems in the \texttt{medicalvalues} software is crucial to provide integration with other systems and datasets.

\begin{figure}[t]
    \centering
    \includegraphics[width=\textwidth]{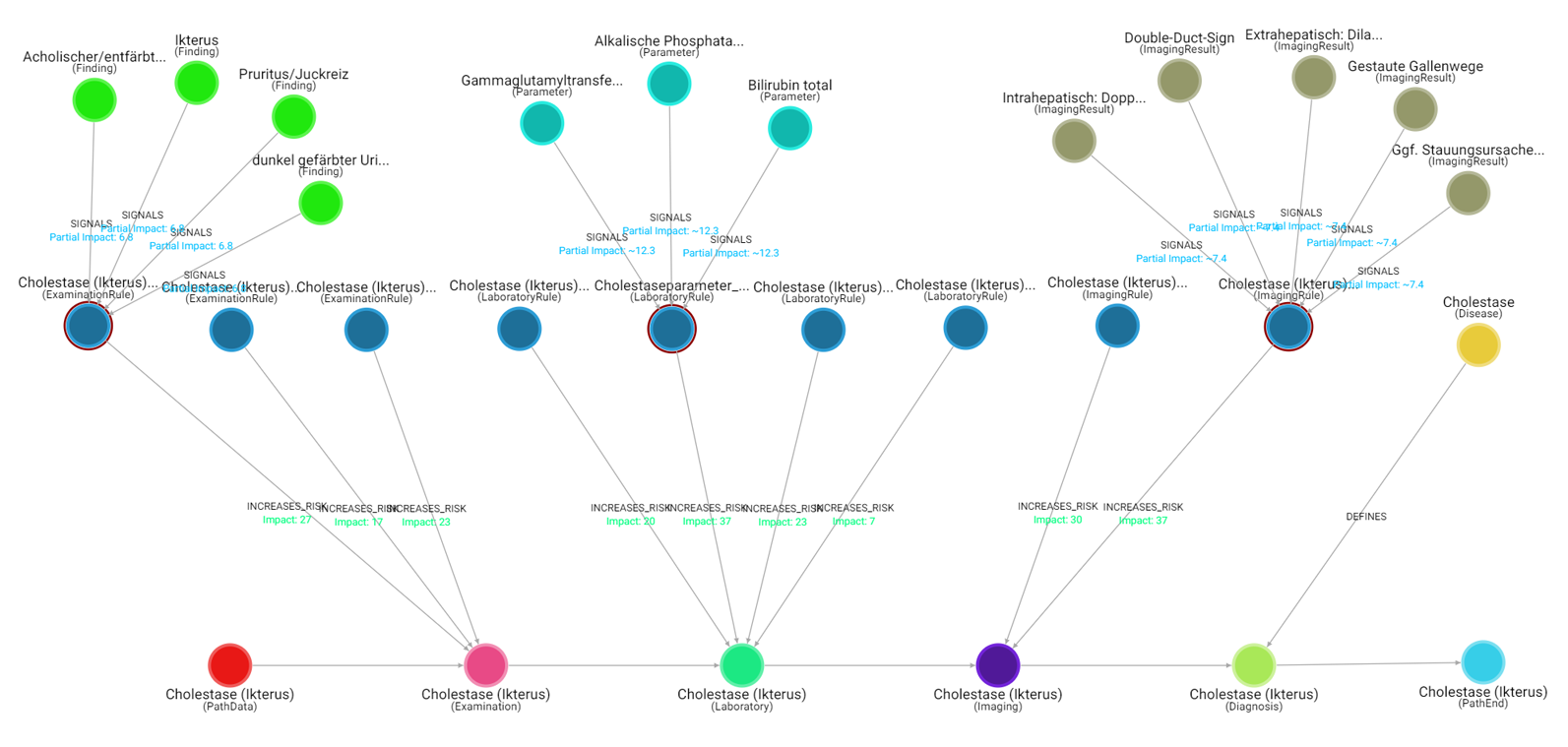}
    \caption{Example diagnosis path for cholestasis, as shown by the \texttt{medicalvalues} knowledge graph editor. The examination of risk factors is followed by a laboratory diagnosis, following the rules via connections to the risk factors. The risk factors presented in the top part of the figure are findings (e.g. symptoms, green nodes), laboratory parameters (turquoise nodes), and imaging results (brown nodes).}
    \label{fig:example_path}
\end{figure}

Figure~\ref{fig:example_path} shows an example diagnosis path for cholestasis. It consists of a first examination of risk factors, which is followed by a laboratory diagnosis. The rules to analyze imaging properties follow the laboratory rules. For every rule type, we show different connections to the risk factors. 
Currently, the \texttt{medicalvalues} knowledge graph contains 411 such diagnosis paths. Each diagnosis path can use different rules: cause rules encode explicitly known causal relationships, examination rules are direct doctor-patient interactions, laboratory rules involve the processing of medical samples in a laboratory, and imaging rules involve the interpretation of medical imaging (e.g., X-ray images). For example, in Fig.~\ref{fig:example_path}, the light green node in the bottom describes a laboratory rule involving the evaluation of three parameters (represented by the three turquoise nodes in the top).

\begin{table}[t]
    \caption{Statistics on the \texttt{medicalvalues} knowledge graph}
    \label{tab:graph_statistics}
    \centering
    \begin{tabular}{lr|lr}
         \multicolumn{2}{c|}{Rules} & \multicolumn{2}{c}{Risk Factors}\\\hline
         ExaminationRule & 1,219 & Finding & 2,240\\
         LaboratoryRule & 871 & Disease & 1,269\\
         ImagingRule & 300 & ImagingResult & 624\\
         ClassifierRule & 158 & Parameter & 1,060\\
         CauseRule & 177 & PathologicalReferenceRange & 171\\
         & & PatientBaseData & 109\\\hline
         \multicolumn{4}{c}{Other node types}\\\hline
         Sample & 48 & Alternative & 36 \\
         ImagingProcedure & 112 & ExternalSystem & 3 \\
         Organ & 23 & ReferenceRange & 1,454 \\
         Source & 1,088 & PathData & 411 \\\hline
        \multicolumn{2}{l}{Total nodes} & \multicolumn{2}{l}{11,373}\\
        \multicolumn{2}{l}{Total edges} & \multicolumn{2}{l}{25,149}\\
    \end{tabular}
\end{table}

In the \texttt{medicalvalues} knowledge graph, there are 2,725 rules which can be used in a diagnosis path, the majority of them being related to examinations (by a physician) and laboratory results. The graph encompasses 1,269 diseases, which can be both an outcome of a diagnosis, as well as a risk factor for others (e.g., Asthma being a risk factor for CoViD-19). In addition, there are 109 artificial patient records, mainly for demonstration and testing purposes. Overall, the graph currently consists of 11,373 nodes and 25,149 edges, as shown in Table~\ref{tab:graph_statistics}. All the knowledge in the graph was curated and validated by medical experts.

\section{Refining Disease Diagnosis Paths with RDF2vec}
\label{sec:refinement}
For our experiments, we first use only the information in the \texttt{medicalvalues} knowledge graph, which contains a set of artificial test patients. 
We then build an enriched graph with additional patient data, extracted from a real dataset (MIMIC-IV), in order to provide extra evidence for refining the original \texttt{medical\-values} knowledge graph.

\subsection{Extending the \texttt{medicalvalues} Knowledge Graph into an Augmented Knowledge Graph}
The patient data is extracted from the Medical Information Mart for Intensive Care IV (MIMIC-IV) database \cite{johnson_alistair_mimic-iv_nodate}. MIMIC-IV consists of data about more than 40,000 patients and is part of the PhysioNet repository of freely-available medical research data \cite{goldberger_physiobank_2000}. The database contains patient data collected at the intensive care units at the Beth Israel Deaconess Medical Center.

We process the data such that we are able to connect patients in MIMIC-IV with diseases of the \texttt{medicalvalues} knowledge graph. Since both MIMIC-IV and the \texttt{medicalvalues} knowledge graph use the same identifiers (i.e., LOINC and ICD-10), the linking can be done directly based on those identifiers. For every patient, the measured laboratory parameters, and the disease, which was recorded at hospital admission, are extracted. Additionally, we store the patients' age and gender. Moreover, the measured parameters are \emph{evaluated}, meaning that a numerical value (e.g., a \texttt{Bilirubin} concentration of 0.43mg/dl in a blood sample) is mapped to an interpreted diagnostic finding, usually on a three-point nominal scale, i.e., stating that the value is \texttt{normal}, \texttt{increased}, or \texttt{decreased}. 

\begin{figure}[t]
    \centering
    \includegraphics[width=0.5\textwidth]{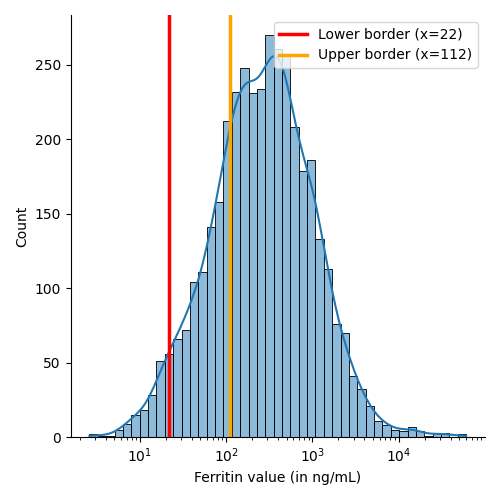}
    \caption{Distribution of \texttt{Ferritin} for adult male patients}
    \label{fig:ferritin_range}
\end{figure}

This categorization is based on the reference ranges of parameters which are defined by the medical experts and coded in the \texttt{medicalvalues} knowledge graph. They usually specify the 0.95-confidence interval of the parameter values of healthy patients. In total, the \texttt{medicalvalues} knowledge graph contains reference ranges for 529 parameters. Figure \ref{fig:ferritin_range} shows the distribution of the values of parameter \texttt{Ferritin} in the MIMIC-IV dataset, together with the corresponding borders of the reference range.\footnote{Please note that the distribution is not representative for the mostly healthy overall population, but contains only hospitalized subjects for which the Ferritin value was determined. This explains why the majority of the sample has a value above the upper bound of the reference range.}

After the processing step, we merge the patients and their evaluated parameters with the diseases in the \texttt{medicalvalues} knowledge graph. As this graph is a labeled property graph, a translation into an RDF graph is needed. This RDF graph contains parameters together with their evaluation as nodes. Every evaluated patient is added to the graph as a new node. The measured parameters are then connected to the parameters of the medical rules.

\begin{figure}[t]
    \centering
    \includegraphics[width=0.7\textwidth]{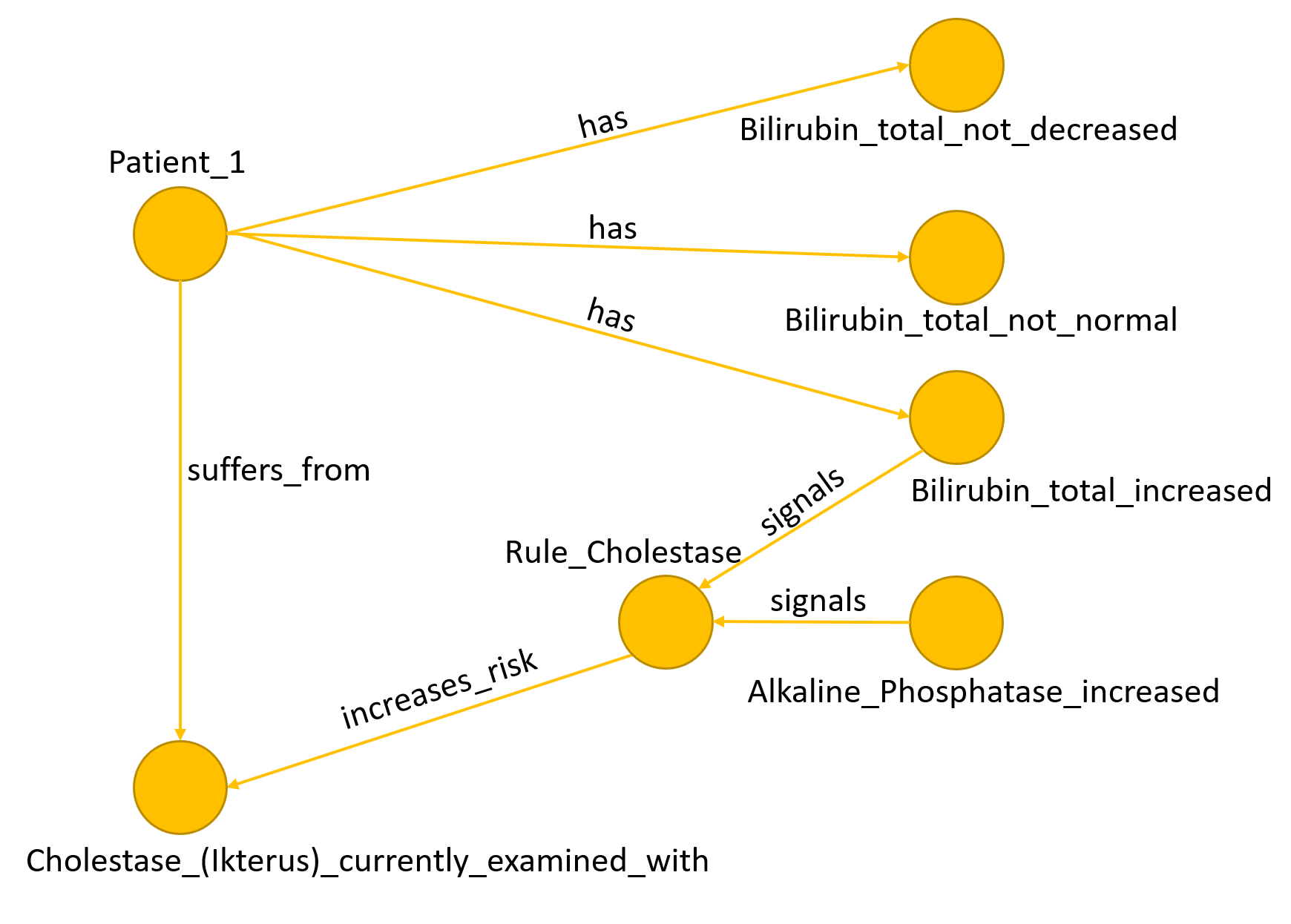}
    \caption{Example of patient data connected to an existing rule in the graph}
    \label{fig:patient_rule}
\end{figure}

Figure~\ref{fig:patient_rule} shows an example of a patient (\texttt{Patient\_1}) connected to a rule. The patient has a recorded disease (\texttt{Cholestase\_(Ikterus)}) and a laboratory parameter (\texttt{Bilirubin\_total\_increased}). That parameter is automatically augmented by additional diagnoses (\texttt{Bilirubin\_ total\_not\_decreased}, \texttt{Bilirubin\_total\_not\_normal}) which are logic consequences of the parameter and simplify the connection of the patient to different diagnostic rules.

The resulting RDF graph has 59,813 nodes and 1,717,685 edges. The average in and out degree of nodes is 28.72.

It is important to point out that the augmented knowledge graph is only used for making predictions, but the patient data are not permanently added to the \texttt{medicalvalues} knowledge graphs. Thereby, privacy issues are avoided.

\subsection{Training RDF2vec Embedding Vectors}
RDF2vec is a method that computes a continuous vector representation for each node in a graph \cite{ristoski2019rdf2vec}. This method consists of two steps: extracting sequences from the graph using random walks, and utilizing the word2vec algorithm to generate vector embeddings from those sequences.

While alternatives to random walks have been proposed as well, the result with those are not yet very conclusive \cite{cochez2017biased,steenwinckel2021walk}, which is why we decided to stick to random walks as a sequence extraction technique. In particular, we compare two flavors of random walks:
\begin{itemize}
    \item \emph{Classic} random walks, as in the original RDF2vec implementation. We start with a fixed number of random walks from each node, only following the outgoing edges. As a result, there is the same number of random walks starting in each node.
    \item \emph{Mid-Walks}, originally introduced for RDF2vec Light \cite{portisch2020rdf2vec}, which also starts a fixed number of random walks per node, but following incoming and outgoing edges. As a result, the nodes can appear in any position in the walk, and this may not necessarily be in the starting position. Mid-walks are assumed to cover a wider variety of knowledge about nodes when not traversing all paths in the graph.
\end{itemize}
In both setups, we use a walk length of 4, and 100 walks per node. We use jRDF2vec to extract the walks and train the embedding vectors.\footnote{\url{https://github.com/dwslab/jRDF2Vec}}

The generated walks are used as input for word2vec using the skip-gram model \cite{mikolov_efficient_2013}.  Word2vec is a neural network that estimates the probabilities of words occurring in the context of other words. The walk sequences can be used as inputs to word2vec, as equivalent to sentences. After the training process the weights of the word2vec network are used as vector embeddings. In our experiments, we consider different dimensionality for the vectors (i.e., 50, 100, 200, 500, 1000, 2000).

\subsection{Predicting Augmentations for Diagnosis Paths}
Given the knowledge encoded in the vector representations, we try to predict links in the knowledge graph. To that end, embedding vectors for each pair of a rule and a risk factor are concatenated and passed to a binary classifier to make a decision whether the risk factor should be included in the corresponding diagnosis rule or not. In figure \ref{fig:patient_rule}, four risk factors (on the right-hand side) and one rule are shown. Therefore, this leads to four possible combinations between the rule and risk factors that can be predicted. The positive samples can be built using the existing relations in the graph. In the example, this is the relation from \textit{Bilirubin\_total\_increased} to \texttt{Rule\_Cholestase} and the relation from \texttt{Alkaline\_Phosphatase\_increased} to \texttt{Rule\_Cholestase}. So, we take the corresponding vector embeddings of rule and risk factor and concatenate them to create a positive sample.

The prediction itself is then modeled as a binary classification task, using the concatenated vectors of the condition and the rule as input, as described in \cite{portisch2021embedding}.

For generating positive samples, the approach of using existing links can be applied. On the other hand, there is no information of negative samples in the \texttt{medicalvalues} knowledge graph. As the graph is currently built and not many disease paths are considered complete, we have to work with the open-world assumption. Therefore, we experiment with three different approaches for building the negative samples:
\begin{itemize}
    \item In the first approach, we randomly sample pairs of rules and risk factors which are not present in the graph, and use them as negative samples (\emph{random}).
    \item We sample random risk factors per rules which are not connected in the graph, and use them as negative samples (\emph{per rule}). This ensures that the classifier sees positive and negative examples for each rule.
    \item In the last approach, we use explicit negations. For example, if we know that \textit{Bilirubin\_total\_increased} is a positive example for the rule \textit{Rule\_Cholestase}, i.e., an increased bilirubin value is a signal for \textit{Cholestase}, we construct a negative example for the \textit{signals} relation using \textit{Bilirubin\_total\_decreased} and the same rule (\emph{explicit}). In other words: for all indicators which have an explicit opposite, such as an increased and a decreased value, we use those opposites for creating negative examples.
\end{itemize}



\begin{figure}[t]
    \centering
    \includegraphics[width=\textwidth]{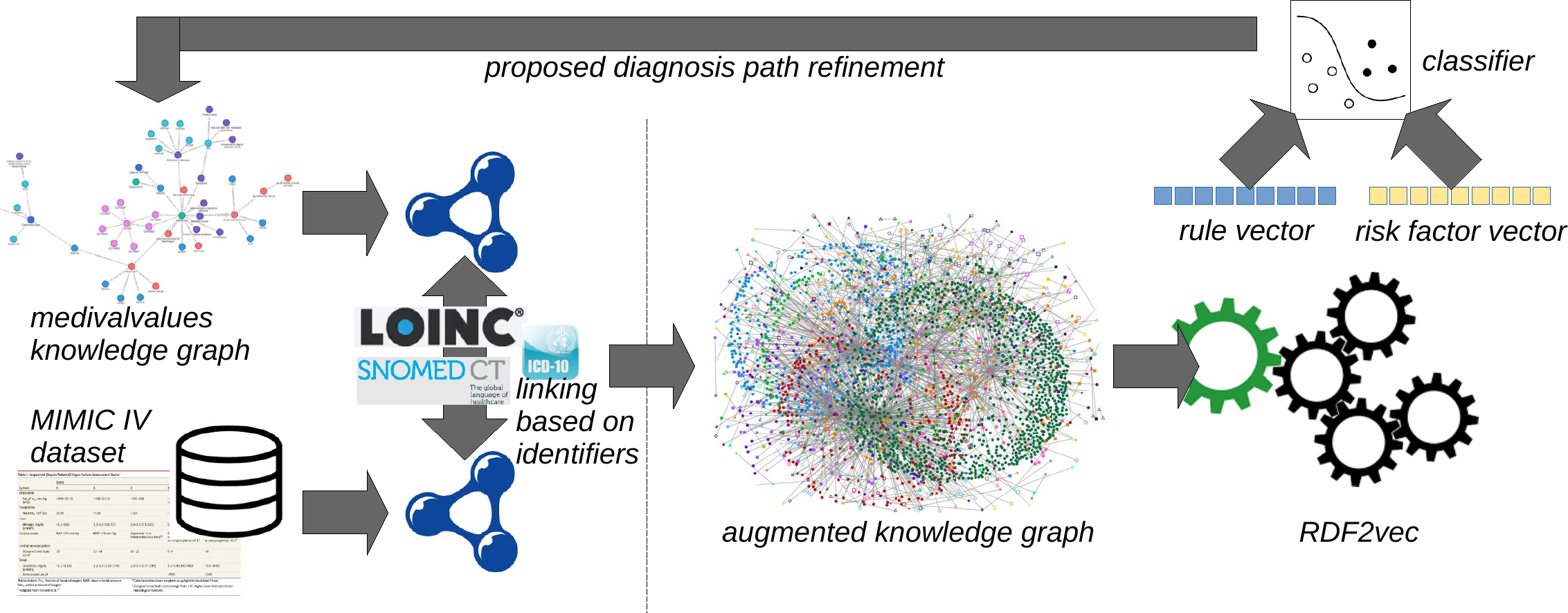}
    \caption{Overall workflow of our approach}
    \label{fig:workflow}
\end{figure}

In our experiments, we use three classifiers, namely Support Vector Machines (SVMs), Logistic Regression, and Random Forests. For all of them, the optimal parameter settings are determined in an internal cross validation in grid search.\footnote{For SVM: kernel function and C, for Logistic Regression: solver and C, for Random Forests: number of trees, number of features, maximum depth, minimum samples in split and leaf nodes, sampling strategy.} It should be pointed out that although the classifiers, as well as the embeddings, are not interpretable, we aim at using them augmenting an interpretable decision system, which, in total, will provide transparent predictions. Fig.~\ref{fig:workflow} shows the overall workflow of our approach.

\section{Evaluation}
\label{sec:evaluation}
The evaluation of the approach is done in two phases. In the first phase, we analyze how well the held-out relations can be reconstructed. We use a train/test split, with tuning via 10-fold cross-validation on the train partition. Here, we perform a number of ablation studies to investigate the influence of different parameters on the model. The code for the experiments is available online.\footnote{\url{https://gitlab.com/medicalvalues-public/mvrdf2vec}}

In the second phase, in order to analyze the model quality manually, we asked a medical expert to review a sample of the produced predictions from the models.

\subsection{Internal Evaluation}
For the internal evaluation, we used the RDF2vec model trained on the merged augmented knowledge graph. In order to measure the impact on using the patient data as additional training data, we also trained an RDF2vec model on the plain \texttt{medicalvalues} knowledge graph as a baseline. We removed 25\% of the relations of a condition and a rule from the graph as a test set. The target of the evaluation is to predict the presence of a relation, given a condition and a rule.

\begin{table}[t]
\caption{Results for the heterogeneous graph including the patient data, contrasted with the \texttt{medicalvalues} knowledge graph alone as a baseline. The table depicts the results for three classifiers (SVM, Logistic Regression (LogReg), and Random Forest (RF) in their respective best performing configuration.}
\label{tab:results_all}
\begin{center}
\begin{tabular}{p{2cm}|r|r|r}
	&	\multicolumn{1}{c|}{LogReg}	&	\multicolumn{1}{c|}{SVM}	&	\multicolumn{1}{c|}{RF}\\
	\hline
\multicolumn{4}{c}{Baseline: \texttt{medicalvalues} graph alone} \\
\hline
Precision	&		0.7199	&	0.7187	&	0.8126\\
Recall	&	0.7385	&	0.7374	&	0.8112\\
F1	&	0.7177	&	0.7285	&	0.8119\\
\hline
\multicolumn{4}{c}{\emph{heterogenous graph}} \\
\hline
Precision	&	0.8718	&	\textbf{0.9079}	&	0.8987\\
Recall	&	0.9189	&	0.9324	&	\textbf{0.9595}\\
F1	&	0.8947	&	0.9200	&	\textbf{0.9281}
\end{tabular}
\end{center}
\end{table} 

Table~\ref{tab:results_all} shows the results of the internal evaluation. 
The F1 score of almost 93\% is very encouraging. Moreover, the results show that using the augmented graph, enriched with information from the patient dataset, drastically improves both the precision and the recall of the prediction.

In addition, we tested the influence of individual methodological decisions of the model in order to better understand their impact. 

Table~\ref{tab:walk_strategies} shows the influence of the two walk strategies (random walks vs. mid-walks). While mid-walks achieve a slightly better results when combined with Random Forests classifier, we decided to stick with random walks since the results are more stable across different classifiers.

\begin{table}[t]
    \caption{Influence of the random walk strategy on the F1 score of the different approaches.}
    \label{tab:walk_strategies}
    \centering
    \begin{tabular}{l|r|r|r||r|r|r}
        & \multicolumn{3}{c||}{Random Walks} & \multicolumn{3}{c}{Mid Walks}\\
        \hline
        & LogReg & SVM & RF & LogReg & SVM & RF \\
        \hline
        Precision	&	0.8718	&	0.9079	&	0.8987	&	0.8800	&	0.8400	&	\textbf{0.9333}\\
        Recall	&	0.9189	&	0.9324	&	\textbf{0.9595}	&	0.8919	&	0.8514	&	0.9459\\
        F1	&	0.8947	&	0.9200	&	0.9281	&	0.8859	&	0.8456	&	\textbf{0.9396}    \\
    \end{tabular}
\end{table}

The negative sampling strategy has also been evaluated. For the  \emph{random} sampling approach, the exact same amount of negatives as positive examples were generated. In the \emph{per rule}  sampling approach, we used different numbers of negatives, leading to differently skewed datasets. The negatives for the \emph{opposite} sampling approach, on the other hand, use a negative example for all positive examples for which a defined opposite exists. We can observe that the best results are obtained using the \emph{opposite} sampling strategy, while adding more negatives (and thereby skewing the dataset) leads to worse results for all classifiers. This shows that the incorporation of domain knowledge for generating negative samples improves the results.

\begin{table}[t]
\caption{F1 scores for different negative label approaches.}
\label{tab:negative_labels}
\centering
\begin{tabular}{l|r|r|r|r|r|r}
 & Random & 1 per rule & 3 per rule & 5 per rule & Opposite \\
 \hline
\# Positive Labels & 200 & 200 & 200 & 200 & 200\\\hline
\# Negative Labels & 200 & 242 & 731 & 1218 & 115\\\hline\hline
LogReg  & 0.9150 & 0.8649 & 0.8000 & 0.6885 & \textbf{0.9404}\\
SVM  & 0.8904 &  0.9028 & 0.8235 & 0.8252 & \textbf{0.9737}\\
Random Forest & 0.9139 & 0.9252 & 0.8966 & 0.8933 & \textbf{0.9733}\\
\end{tabular}
\end{table} 

In a final ablation study, we explore the embedding dimension, as shown in Table~\ref{tab:vector_dimension}. While the original RDF2vec papers typically used 200 or 500 dimensions \cite{ristoski2019rdf2vec}, and studies on other knowledge graph showed good results also for a smaller number of dimensions \cite{portisch2020rdf2vec}, we observed the best results for 1,000 and 2,000 dimensions.
Ultimately, the best results were obtained by using opposites for negative labels, 1,000 dimensions, and a random forest classifier.
\begin{table}[t]
    \caption{F1 scores for different vector dimensions.}
    \label{tab:vector_dimension}
    \centering
    \begin{tabular}{l|r|r|r|r|r|r} 
 & 50 & 100 & 200 & 500 & 1,000 & 2,000\\\hline\hline
LogReg  & 0.8462 & 0.8718 & 0.8947 & 0.8947 & 0.9189 & \textbf{0.9388} \\
SVM  & 0.8903 &  0.9211 & 0.9200 & 0.9067 & \textbf{0.9306} & 0.8933\\
Random Forest  & 0.8874 & 0.9221 & 0.9281 & 0.9467 & \textbf{0.9517} & 0.9315\\
    \end{tabular}
\end{table} 

Besides the qualitative exploration, we also study the computational performance. All experiments were performed on a standard commodity laptop. The most time-consuming step was the conversion of the MIMIC-IV database and its conversion to a knowledge graph, which took more than 12 hours. On the other hand, training the RDF2vec model and computing the predictions could be performed in under 10 minutes, even for the higher dimensionalities.

\subsection{External Evaluation}
In order to harden the results in a more realistic setup, we presented a set of predictions to a medical domain expert. We used Random Forests with \emph{explicit} negative sampling that was the best performing setup, and we collected predictions for pairs of rules and parameters from the MIMIC-IV dataset, where for the latter, we made the restriction that at least five evaluations (i.e., mappings to a nominal interpretation, see above) exist. The rules referred to a set of selected diseases. These diseases were chosen with help of the medical expert with the criterion that the disease paths includes at least one of the extracted and successfully evaluated parameters from MIMIC-IV. In total, 318 predictions were created that way and shown to the expert, who was asked to rate them on a five-point scale, as shown in Table~\ref{tab:codes_expert}.

\begin{table}[t]
    \caption{Codes for the expert-driven evaluation}
    \label{tab:codes_expert}
    \centering
    \begin{tabular}{r|l}
         Code &  Description \\
         \hline
        1 & The relation clearly exists. \\
        2 & It is not clear whether the relation exists, but it is plausible. \\
        3 & It is not clear whether the relation exists, but the chance is low. \\
        4 & The relation clearly does not exist. \\
        5 & No statement about the relation can be made.    \end{tabular}
\end{table}

Overall, we evaluated new relations for six diseases, i.e., \textit{iron deficiency anemia (IDA)}, \textit{Hypothyroidism (Hypo)}, \textit{autoimmune hepatitis (AH)}, \textit{nonautoimmune hemolytic anemias (NHA)}, \textit{Benign neoplasm of pituitary gland (BNPG)}, and the \textit{tumor lysis syndrome (TLS)}. The results are shown in Table~\ref{tab:expert_results}. It can be observed that the number of correct (37) and plausible (24) predictions is clearly higher than that of unlikely (11) and wrong (2) ones. The majority, however, are relations for which no statement can be made.

Examining some of the predictions in detail, we observe that only a small amount of contrary predictions (increased and decreased for the same parameter) was made. E.g., for \textit{hypothyroidism}, only for two parameters (\textit{IgA}, \textit{Cholesterol}) a relation for both an \emph{increased} and a \emph{decreased} value is predicted; however, the relation between the disease and the value was correctly identified, and an expert can easily discard the wrong condition.

The predictions for \textit{iron deficiency anemia} also contain an interesting result. The only two predictions marked as wrong by the expert were made for the parameter \textit{Ferritin}. A decreased \textit{Ferritin} value is a key indicator for \textit{iron deficiency anemia}, but the model predicted the opposite. To understand the prediction, we looked at the records of the patients suffering from \textit{iron deficiency anemia}. In total, 3,529 patients were associated with \textit{iron deficiency anemia}, and Ferriting was measured only for 1,423 of them. 
This could hint at the fact that the disease was already known, and the patients were in the hospital for different reasons. 
This indicates that the prediction itself was not necessarily wrong, but based on a wrong assumption, i.e., the model treating the diagnosed disease and pre-known (maybe even not acute) diseases alike.

Summarizing, the evaluation shows two key outcomes: (1) it is possible to produce novel relations to present an expert with, e.g., for augmenting the knowledge graph in a human-in-the-loop setting, and (2) additional evidence and information is required in many cases to make an informed decision, i.e., the prediction alone is not sufficient.

\section{Conclusion and Outlook}
\label{sec:conclusion}
In this paper, we have introduced the \texttt{medicalvalues} knowledge graph, which is used for medical diagnosis using so-called diagnosis paths. Those paths allow for a transparent prediction of a patient's disease. Since the paths are developed manually, they are notoriously incomplete.

To tackle this incompleteness, we have introduced an approach which first enriches the \texttt{medicalvalues} knowledge graph into a augmented graph, connecting it to a large dataset of patient records. On that augmented graph, we have trained vector embeddings with RDF2vec, which are used to predict completions of the existing diagnosis paths. Both in an internal validation as well as in an expert evaluation, we have shown that the prediction of such extensions is possible with high precision. This methodology of enriching the graph and producing predictions therewith is independent of the task and domain at hand.

One key limitation of the approach is the external data used, which is data gathered from intensive care units. Therefore, diseases which do rarely lead to treatments in intensive care are not well covered. In order to augment diagnosis paths for as versatile diseases as possible, other external datasets should be considered as well. Here, the connectors to clinic information systems (CIS) and laboratory information systems (LIS) may also add large-scale instance data in the future, which can also be exploited with the same methodology.

So far, drugs are not represented in the \texttt{medicalvalues} knowledge graph. In the future, we would like to include them, both as a part of a patient's medical history (i.e., existing medication), as well as possible treatments once a diagnosis is made. To that end, we plan to augment the graph with existing datasets on drugs and drug interactions.

When contrasting the internal and external evaluation, we have seen that while in both cases, the precision is rather high, there are still differences, most prominently the high number of predictions which an expert cannot make any statement on. Here, it would be interesting which kind of additional information the expert needs to make a decision. Moreover, we plan for a study with a larger pool of medical experts from different medical sub fields in order to better assess the capabilities of the approach in different medical fields.

A lightweight augmentation would be pointing the expert to the original patient files as evidence. A more complex approach could incorporate search in scientific databases like PubMed \cite{mcentyre2001pubmed} for articles which contain both the disease and the symptom at hand. 

\begin{table}[t]
    \caption{Outcome of the expert evaluation}
    \label{tab:expert_results}
    \centering
    \begin{tabular}{r|r|r|r|r|r|r|r}
        & IDA & Hypo & AH & NHA & BNPG & TLS &  Total\\
        \hline
        ICD-10 & D50.9 & E03.9 & K75.4 & D59.4 & D35.2 &  E88.3 & - \\
        Occurrences & 3,530 & 51,400 & 669 & 63 & 324 & 97 & 56,083\\ 
        \hline
        1 & 4 & 8 & 5 & 7 &  1 & 12 &  37 \\
        2 & 0 & 4 & 5 & 2 & 0 & 13 &  24 \\
        3 & 3 & 2 & 4 & 0 & 0 & 2 &  11 \\
        4 & 2 & 0 & 0 & 0 & 0  & 0 &  2 \\
        5 & 45 & 39 & 39 & 44 & 52 & 26 &  244 
    \end{tabular}
\end{table}

In the future, we want to explore the utility of other embedding models beyond RDF2vec \cite{portisch2021embedding}. From a technical perspective, other solutions are also possible. For example, explanation models \cite{zhang2019interaction} or symbolic rule learners \cite{meilicke2019anytime} could be used in addition to the prediction method in order to provide the expert with more fine-grained explanations for the prediction instead of just the indication of a missing condition itself.

Once such explanation models and/or additional clues are available, we plan to have a larger expert evaluation study, which will not only incorporate a larger number of medical experts, but also a contrastive evaluation of which external clues (e.g., patient records, scientific sources, etc.) is considered the most helpful by the medical experts.

In the medical knowledge graph used in this paper, it is also possible to incorporate edge weights (reflecting, e.g., the frequency at which a particular influence of a condition on a disease is observed). Such weights could be incorporated in the embeddings \cite{taweel2020towards} in order to further improve the results (and, in particular, give more emphasis on more prevalent relations). Moreover, numeric values are currently only utilized via discretization, and we will explore more sophisticated approaches in the future.

In summary, we have shown the potential use of knowledge graphs in medical diagnosis systems, and discussed how knowledge graph embeddings can be utilized to refine those knowledge graphs, while still retaining the explainability and accountability of the overall systems. The method of combining the knowledge graph to be augmented with auxiliary data into an augmented graph has been shown to be an effective means of using external data for knowledge graph refinement, which can also be transferred to other domains.

\bibliographystyle{splncs04}
\bibliography{bibliography}

\end{document}